\pdfoutput=1
\PassOptionsToPackage{table, dvipsnames}{xcolor}
\documentclass[11pt]{article}

\usepackage[final]{acl}

\usepackage{times}
\usepackage{latexsym}

\usepackage[T1]{fontenc}
\usepackage[utf8]{inputenc}

\usepackage{microtype}

\usepackage{inconsolata}

\usepackage{graphicx}

\title{R.R.: Unveiling LLM Training Privacy through Recollection and Ranking}

\author{%
\textbf{Wenlong Meng}\textsuperscript{1},
\textbf{Zhenyuan Guo}\textsuperscript{1}, 
\textbf{Lenan Wu}\textsuperscript{1}, 
\textbf{Chen Gong}\textsuperscript{2},
\textbf{Wenyan Liu}\textsuperscript{3}, \\
\textbf{Weixian Li}\textsuperscript{3},
\textbf{Chengkun Wei}\textsuperscript{1}$^{\dag}$, 
\textbf{Wenzhi Chen}\textsuperscript{1}
\\
$^1$Zhejiang University, $^2$University of Virginia, $^3$Ant Group \\
\texttt{\{mengwl, zhenyuanguo, weichengkun, chenwz\}@zju.edu.cn}
}

\usepackage[noend]{algpseudocode}
\usepackage{amssymb}
\usepackage{amsthm}
\usepackage{tikz}
\usepackage{amsmath, amsfonts}
\usepackage{mathtools}
\usepackage{graphicx}
\usepackage{textcomp} % special symbols
\usepackage[T1]{fontenc}
\usepackage[utf8]{inputenc}
\usepackage{xspace}
\usepackage{multirow}
\usepackage{diagbox}
\usepackage{booktabs}
\usepackage{caption}
\usepackage{subcaption}
\usepackage{float}
\usepackage{makecell}
\usepackage{paralist}
\usepackage{enumitem}
\usepackage[skins]{tcolorbox}
\usepackage[ruled,lined,linesnumbered]{algorithm2e}
\usepackage{algorithmicx}
\usepackage{pifont}
\usepackage[pagebackref=true,breaklinks=true,colorlinks,bookmarks=false]{}
\usepackage{hyperref}
\hypersetup{
  colorlinks,
  linkcolor={PineGreen},
  citecolor={RoyalBlue},
  urlcolor={Bittersweet}
}
\definecolor{mycolor}{RGB}{246, 196, 188}
\definecolor{codegray}{rgb}{0.4, 0.4, 0.4}
\definecolor{cautioncolor}{RGB}{255,0,0}
\usepackage{soul}
\sethlcolor{mycolor}

\colorlet{lightgray}{gray!20}

\SetCommentSty{mycommfont}

\newcommand{\appref}[1]{\hyperref[#1]{Appendix~\ref*{#1}}}

\definecolor{redbg}{RGB}{254,241,240}
\definecolor{redoutline}{RGB}{252,163,152}
\definecolor{redtext}{RGB}{207,24,34}
\DeclareRobustCommand*{\rectangled}[1]{%
  \tikz[baseline=(char.base)]\node[anchor=south west, draw, rectangle, thick, rounded corners=0.2mm, inner sep=2pt, fill=redbg, draw=redoutline,text=redtext](char){#1} ;}

\newcommand{\mypara}[1]{\smallskip \noindent\textbf{#1.} \xspace}
\newcommand{\myparatight}[1]{\noindent\textbf{#1.} \xspace}

\newcommand{\method}{\ensuremath{\mathsf{R.R.}}\xspace}

\newcommand{\spbase}{{P2P}\xspace}

\newcommand{\block}[1]{%
  \raisebox{\dimexpr(\fontcharht\font`X-1em)/2}{\rule{0.5em}{#1\dimexpr1em/8}}%
}
\DeclareUnicodeCharacter{2581}{\block{1}}
\DeclareUnicodeCharacter{2582}{\block{2}}
\DeclareUnicodeCharacter{2583}{\block{3}}
\DeclareUnicodeCharacter{2584}{\block{4}}
\DeclareUnicodeCharacter{2585}{\block{5}}
\DeclareUnicodeCharacter{2586}{\block{6}}
\DeclareUnicodeCharacter{2587}{\block{7}}
\DeclareUnicodeCharacter{2588}{\block{8}}

\newtcolorbox{mybox}[2][]{text width=\textwidth, fontupper=\normalsize,
fonttitle=\bfseries\sffamily\normalsize, colbacktitle=codegray, enhanced,
boxed title style={sharp corners}, top=4pt, bottom=2pt, left=2pt, right=2pt,
title=#2, colback=gray!10}

\theoremstyle{plain}
\newtheorem{theorem}{Theorem}[section]

\theoremstyle{definition}

\theoremstyle{remark}

\usepackage{etoolbox} % For \patchcmd

\makeatletter
\patchcmd{\algocf@makecaption@ruled}{\hsize}{\textwidth}{}{} % Caption to stretch full text width
\patchcmd{\@algocf@start}{-1.5em}{0em}{}{} % For // to right margin
\makeatother

\begin{document}
\maketitle
\begin{NoHyper}
\def\thefootnote{\dag}\footnotetext{Corresponding author}
\end{NoHyper}

\begin{abstract}
Large Language Models (LLMs) pose significant privacy risks, potentially leaking training data due to implicit memorization. Existing privacy attacks primarily focus on membership inference attacks (MIAs) or data extraction attacks, but reconstructing specific personally identifiable information (PII) in LLMs' training data remains challenging. In this paper, we propose \method (Recollect and Rank), a novel two-step privacy stealing attack that enables attackers to reconstruct PII entities from scrubbed training data where the PII entities have been masked. In the first stage, we introduce a prompt paradigm named recollection, which instructs the LLM to repeat a masked text but fill in masks. Then we can use PII identifiers to extract recollected PII candidates. In the second stage, we design a new criterion to score each PII candidate and rank them. Motivated by membership inference, we leverage the reference model as a calibration to our criterion. Experiments across three popular PII datasets demonstrate that the \method achieves better PII identification performance than baselines. These results highlight the vulnerability of LLMs to PII leakage even when training data has been scrubbed. We release our code and datasets at GitHub.\footnote{\url{https://github.com/meng-wenlong/RR}}

\end{abstract}

\section{Introduction}

Large Language Models (LLMs) have been widely adopted in various applications, like automated content generation~\citep{girotra2023ideas, roziere2023code} to personalized virtual assistants~\citep{nam2024using, shao2024deepseekmath}.
This success is driven by massive training datasets, which raise significant concerns about privacy leakage.
Previous research has shown that LLMs can memorize portions of their training data, even before overfitting~\citep{carlini2019secret}. These training datasets of LLM may contain sensitive personally identifiable information (PII), such as names, phone numbers, and email addresses. Additionally, LLMs are usually fine-tuned with custom datasets to meet application needs, which may introduce further privacy risks.

\begin{figure}
    \centering
    \includegraphics[width=\columnwidth]{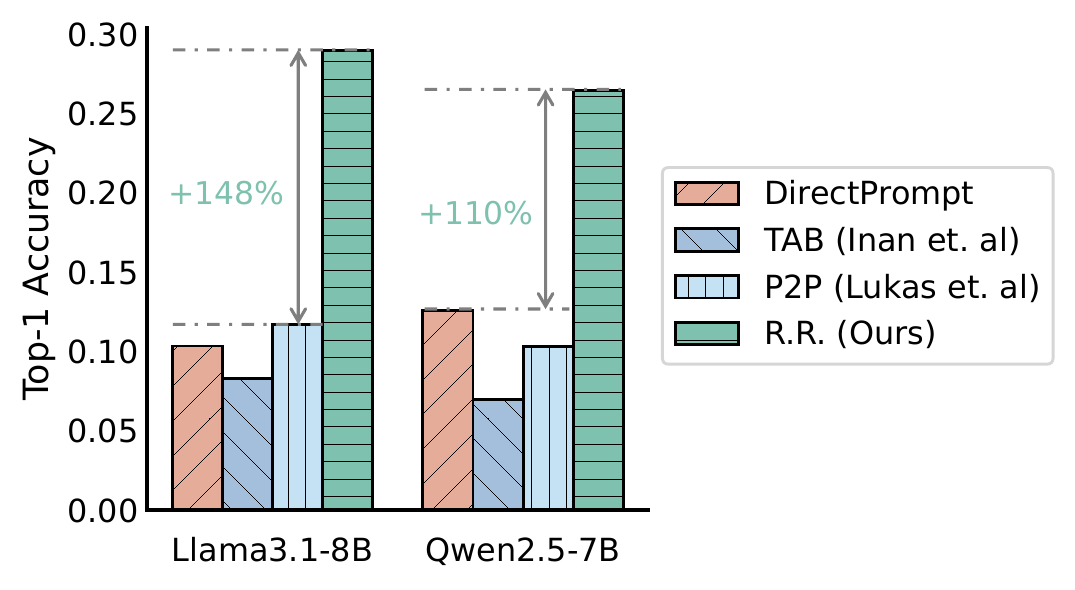}
    \caption{Comparison of PII prediction accuracies on NeurIPS LLM-PC dataset. \method achieves an improvement of over 100\% compared to previous SOTAs.}
    \label{fig:intro}
\end{figure}

LLM memorization is an inherent and inevitable part of training. Thus, the security of LLM privacy is a crucial and widespread topic of public concern. This paper focuses on attacks that steal the privacy of training datasets.
One widely studied privacy attack is \textit{membership inference attack} (MIA)~\citep{mireshghallah2022quantifying, fu2024membership}, which aims to determine whether a specific data record was included in the training set.
However, MIAs targeting LLMs usually require the attacker to obtain complete text samples, which is often impractical in real-world scenarios~\citep{mattern2023membership, duan2024membership}.
Another line of attack is \textit{data extraction}~\citep{carlini2021extracting, mireshghallah2022empirical}, where an adversary attempts to reconstruct training data as much as possible.
However, the impact of data extraction is limited, as the extracted data may not contain the target information an attacker seeks.

Recently, \citealp{lukas2023analyzing} proposed \textit{PII reconstruction attack}, which refers to the task where an attacker attempts to recover the masked PII entities in a piece of text.
For example, in the masked sentence ``The shipment for \rectangled{\texttt{[MASK]}} was delivered to 1234 Elm Street,'' if the attacker successfully reconstructs \rectangled{\texttt{[MASK]}}, they can identify the individual residing at that address. This practical attack has attracted great attention in the community.
Scrubbed datasets publicly released by LLM developers may serve as potential targets for such attacks.
Nevertheless, existing PII reconstruction methods based on prefix continuation~\citep{inan2021training} or perplexity scoring~\citep{lukas2023analyzing}, lack contextual dependencies and insights from the original LLM, yielding a reconstruction accuracy below 10\%.

\mypara{Our Contributions}
In this paper, we propose \method (\underline{R}ecollect and \underline{R}ank), a novel and effective PII reconstruction method that assumes black-box access to the victim LLM.
\method consists of two stages: (1) recollection, generating PII candidates, and (2) ranking, identifying the candidate most likely to be PII from candidates. 

In the recollection stage, we provide the entire masked text as input to the victim LLM and prompt it to reproduce the text without masks.
Unlike traditional prefix continuation methods, which rely solely on the preceding text, our approach leverages both preceding and succeeding contexts.
Once the model finished generating the recalled text, \method uses a PII identifier to extract internal PII entities.
After repeating recollection several times, we get a PII candidates pool.
In the ranking stage, extracted PII candidates are inserted into the mask placeholder. 
We use the cross-entropy loss as the criterion for ranking, and the candidate with the lowest loss is selected as the reconstructed result.
Inspired by reference-based MIAs~\citep{ye2022enhanced, zeng2024exploringmemorizationfinetunedlanguage}, we use the loss from the pre-trained model, on which the victim model is fine-tuned, as a reference to calibrate the ranking criterion.
Interestingly, we observe that reference calibration does not always improve ranking accuracy. 
To address it, we propose a biased ranking criterion that blends uncalibrated and calibrated criteria.
We theoretically prove that our proposed criterion preserves the strengths of both.

Experimental results of four LLMs show that \method achieves an average top-1 accuracy of 25.73\% on the NeurIPS LLM-PC dataset~\citep{li2024the}, marking a more than 100\% improvement over previous SOTAs, as presented in \autoref{fig:intro}.
In summary, our contributions are as follows:

\begin{itemize}
    \item We propose a novel PII reconstruction attack named \method which leverages recollection to generate PII candidates and employs a biased cross-entropy loss for ranking. This paper shows that \textit{scrubbed training data is not safe}, calling for increased attention to this threat.
    \item We take the first step in using reference calibration for PII ranking and propose a new criterion that combines calibrated and uncalibrated criteria.
    \item Extensive experiments across three PII datasets and four popular LLMs show that R.R. achieves an average 122\% improvement in top-1 accuracy compared to baselines.
\end{itemize}

\begin{figure*}[!t]
    \centering
    \includegraphics[width=\textwidth]{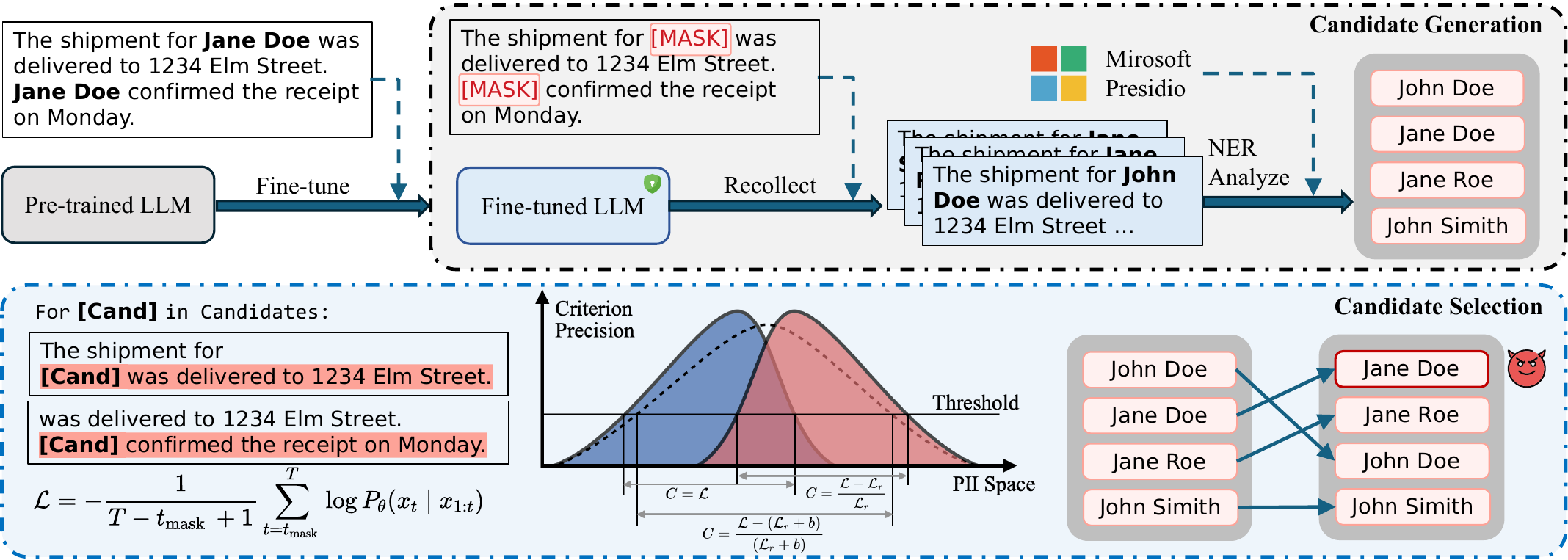}
    \caption{Overview of \method. \method has two steps: candidate generation and selection. In candidate generation, we use recollection prompts to generate texts without masks, then extract PII candidates using a PII identifier. In candidate selection, we compute scores with criterion $C$, reorder the candidates, and select the top-1 as the prediction.}
    \label{fig:overview}
\end{figure*}

\section{Preliminaries}

\subsection{LLM Training and PII}

\mypara{LLM Training}
We train LLMs using next-token prediction, where the model learns to maximize the probability of correctly predicting the next token given a sequence of preceding tokens.
The objective is formulated to minimize cross-entropy loss.
Given a sequence of tokens $X=\left(x_1, x_2, \ldots, x_T\right)$, the cross-entropy loss is defined as:
\begin{equation}\label{eq:loss}
\mathcal{L}=-\frac{1}{T}\sum_{t=1}^T \log P_\theta\left(x_t \mid x_{1:t-1}\right),
\end{equation}
where $P_\theta\left(x_t \mid x_{1:t-1}\right)$ denotes the probability assigned by LLM with parameter $\theta$.

\mypara{Personally Identifiable Information (PII)}
PII refers to any data that can be used to uniquely identify an individual, such as names, addresses, phone numbers, or social security numbers.
Since LLMs are often trained on vast real-world texts, they inevitably ingest and memorize a significant amount of PII.
To detect PII in natural language text, Named Entity Recognition (NER) models are commonly employed. NER is a token classification task, where each token in a sequence is assigned a label corresponding to a specific entity type.
Several industry-grade PII detection services utilize NER-based models to identify and redact sensitive information~\citep{MsPresidio,aws_comprehend}.

\subsection{PII Leakage Attacks}

\mypara{Attacks Taxonomy}
According to~\citealp{lukas2023analyzing} proposed, PII leakage attacks can be categorized into three types based on the adversary's capabilities: PII extraction, reconstruction, and inference attacks.
Extraction attackers have no auxiliary information, while reconstruction attackers have knowledge of masked training texts (e.g., scrubbed datasets) and aim to recover the missing PII.
Inference attacks go a step further, where the attacker possesses both masked training texts and PII candidates.
The goal of inference is to associate a piece of PII with a given context.

\mypara{Threat Model}
This paper focuses on PII reconstruction attacks because obtaining masked training data is practical in real-world scenarios.
%and such data allows attackers to exploit contextual dependencies to recover masked PII entities.
We assume an adversary with black-box API access to the victim LLM, allowing them to request a continuation or query next-token probabilities to compute cross-entropy loss and PPL.
Since API calls entail monetary costs, the adversary should minimize request or query times.
% Experimental results (Section 4.4) prove that Our R.R. 在 candidates generation 阶段达到同样的 PII Recall 需要更少的 query 次数. In candidates selection stage, we 计算loss的时候忽略PII前的content，减少了计算loss的所需的query次数。
Experimental results (\autoref{sec:ablation_studies}) prove that our \method requires fewer queries to achieve the same PII recall in candidate generation.
In candidate selection, we neglect the content preceding PII when computing loss, reducing the query count.
Additionally, we assume the adversary knows the reference model, i.e., the base model on which the victim LLM is fine-tuned.
It is a reasonable assumption since modern LLM development is usually based on open-source LLMs.
And according to popular open-source LLM licenses, such as OpenRAIL-M~\citep{openrail-m} and Apache-2.0~\cite{apache2}, the developer needs to manifest the source when employing a modification of the LLM for commercial use.
Even if the attacker uses an incorrect reference model, we show that it causes a small degradation in reconstruction accuracy (\autoref{app:incorrect_reference}).

\section{\method Methodology}

The overall pipeline of \method is illustrated in~\autoref{fig:overview}. Our \method reconstructs masked PII in two steps: candidate generation and candidate selection. 
During candidate generation, \method iteratively induces the victim LLM to generate possible PII candidates, which are then ranked in candidate selection.
The top-1 candidate is selected as the reconstructed result.

\subsection{Problem Statement}

Our attack involves two participants: an LLM developer and an attacker.
The LLM developer fine-tunes a publicly available pre-trained LLM, denoted as $\pi_\theta$, on a private dataset $\mathcal{D}_p=\left\{X_i \mid i=1,2, \ldots\right\}$, obtaining a fine-tuned LLM $\pi_{\theta^\prime}$.
The developer then deploys the custom LLM on the cloud and provides a query API.
The attacker possesses several masked texts, denoted as $\mathcal{D}_a=\left\{\widetilde{X}_i \mid i=1,2, \ldots\right\}$, where each masked text takes the form:
\begin{equation}
    \widetilde{X}_i=\left(x_{i, 1}, x_{i, 2}, \ldots, \rectangled{\texttt{[MASK]}}, \ldots, x_T\right).
\end{equation}
For each $\widetilde{X}_i \in \mathcal{D}_a$, there exists a sequence $M=\left(m_i\right)_{i=1}^k$ such that%\gcc{Why do we need M? please explain it here.}:
\begin{equation}
    X_i=\widetilde{X}_i \circ M \in \mathcal{D}_p,
\end{equation}
where $\circ$ represents the operation of replacing \rectangled{\texttt{[MASK]}} in $\widetilde{X}_i$ with $M$. The attacker's goal is to recover $M$.
For brevity, we use $\mathcal{L}(M)$ to represent $\mathcal{L}(\widetilde{X} \circ M)$ in the rest of this paper.

In practice, $\widetilde{X}_i$ may contain multiple \rectangled{\texttt{[MASK]}}s, which indicate entity types related to PII, such as \rectangled{\texttt{[DATE\_TIME]}}.
The entity types considered in this paper are listed in~\autoref{tab:pii_types} of~\autoref{app:dataset_details}.

\subsection{Candidate Generation}

This step uses designed prompts to induce the victim LLM to output memorized PII.
We construct prompts using a recollection approach, where the masked text $\widetilde{X}_i$ is directly fed into the LLM, prompting it to repeat the given text while omitting the masked tokens.
We then use Microsoft Presidio to extract PII of specific types from the recollected text. 
Our motivation is to leverage all available information as much as possible.
Through the recollection paradigm, we could leverage all contexts the attacker has.

The most widely used previous candidate generation method is TAB~\citep{inan2021training}, which generates completions using the true prefix. TAB behaves like the Tab autocompletion method which automatically fills in table entries, mimicking the behavior of the Tab key.
However, its major limitation is that it cannot leverage information from the text following the \rectangled{\texttt{[MASK]}}.
For instance, in the sentence: ``The shipment for \rectangled{\texttt{[MASK]}} was delivered to 1234 Elm Street.'' TAB fails to utilize the address ``1234 Elm Street'' to infer the missing PII.
Additionally, if other masks appear in the preceding text, TAB requires truncation to prevent the LLM from generating mask tokens.
For example, if ``shipment'' is also masked in the above sentence, TAB can only use ``for'' as the continuation prefix.
\citealp{lukas2023analyzing} addressed this issue by filling in previous masks using a \textit{Masked Language Model} (MLM).
However, this approach assumes that the token length of the masked PII is known in advance, which is often impractical in real-world scenarios.
Conversely, recollection maximizes the use of both preceding and following context while aligning with the LLMs' training paradigm.

\subsection{Candidate Selection}
\label{sec:candidate_selection}

Assume that, during the candidate generation step, \method produces $n$ PII candidates $\mathcal{M}=\left\{M_i\right\}_{i=1}^n$ for a given mask.
In this step, we aim to assign a score, which indicates fitness to the mask, to each $M_i \in \mathcal{M}$. Then, we rank them and select the top-1 candidate as the final prediction.
In short, our ranking criterion is:
\begin{equation}\label{eq:final_c}
    C\left(M_i\right)=\frac{\mathcal{L}^p\left(M_i\right)-\left(\mathcal{L}_r^p\left(M_i\right)+b\right)}{\mathcal{L}_r^p\left(M_i\right)+b},
\end{equation}
where $\mathcal{L}^p$ and $\mathcal{L}^p_r$ represent the \textit{partial cross-entropy loss} in the victim model and the reference model, respectively.
$b$ is a constant positive number specific to each LLM.

\mypara{Partial Cross-Entropy Loss} We use cross-entropy loss as the base criterion because it allows us to easily sum multiple criteria in cases where a PII mask appears multiple times.
As shown in~\autoref{fig:overview}, when a PII mask appears multiple times in $X_i$, our proposed \method employs a greedy strategy that divides $X_i$ into segments at each PII mask location, computes the score for each segment, and then sums the scores.
In this way, we reduce the computational complexity of loss calculation from $\mathcal{O}\left( {n_c}^{n_t} \right)$ to $\mathcal{O}\left( n_c \cdot n_t \right)$, where $n_c$ and $n_t$ are the number of candidates for reach PII type and the number of PII types, respectively.
Since most LLMs use unidirectional attention, the prediction probabilities for tokens preceding a PII mask will not change due to variations in $M_i$.
Therefore, when calculating the cross-entropy loss, we only consider the text from $M_i$ and onwards,
\begin{equation}\label{eq:partial_loss}
    \mathcal{L}^p=-\frac{\sum_{t=t_{\text {mask }}}^T \log P_\theta\left(x_t \mid x_{1:t}\right)}{T-t_{mask}+1},
\end{equation}
where $t_{mask}$ is the position of the mask token.

The effect of the preceding text on cross-entropy loss can be disregarded when a PII mask appears only once.
However, when the PII mask appears multiple times, the varying lengths of the clauses can cause the normalization factor $\frac{1}{T}$ in~\autoref{eq:loss} to diminish the contribution of longer clauses.
Since $\mathcal{L}^p$ more accurately reflects the impact of inserting $M_i$ in most cases, we use $\mathcal{L}$ to represent the partial cross-entropy loss in the rest of this paper.

\mypara{Revising Biased Reference Calibration}
The pre-trained model has already been trained on some texts.
Even if the victim model's fine-tuning private dataset does not include those texts, their loss may still be lower than private texts.
Actually, the victim model is trained on two datasets, (1) the pretraining dataset and (2) the fine-tuning private dataset.
The overlap between the pre-training and fine-tuning datasets reduces the accuracy of the attack on the private fine-tuning dataset. This challenge motivates us to adjust~\autoref{eq:partial_loss} to avoid confusion between these two datasets and exclude the pretraining dataset's terrible impact.

In MIA, researchers use reference models (i.e., the pre-trained models) to adjust the criteria used for determining whether a text is included in the private training data.
Inspired by reference-based MIAs, we propose reference calibration into the PII inference attack.
To our knowledge, we are the first to do so.
The most commonly used calibration method in MIA is to measure the ratio of the decrease in $\mathcal{L}$ relative to $\mathcal{L}_r$, that is,
\begin{equation}
    C_r = \frac{\mathcal{L} - \mathcal{L}_r}{\mathcal{L}_r}.
\end{equation}
This approach helps to refine the criterion by accounting for the reference model's influence.
However, we empirically find that \textit{reference-based criterion is not always superior to simple loss; rather, they perform comparably in most cases.}

From our investigation, The PII in the training dataset correctly identified using $\mathcal{L}$ for calibration is different from that identified using $\mathcal{C}_r$.
For example, when reconstructing the LLM-PC dataset on a fine-tuned Llama3.1-8B model, $\mathcal{L}$ correctly predicts 1,627 PII entities, while $C_r$ correctly predicts 1,689 PII entities.
Among these, 132 PII entities are correctly predicted by $\mathcal{L}$ but incorrectly predicted by $C_r$, and 194 PII entities are correctly predicted by $C_r$ but incorrectly predicted by $\mathcal{L}$.
These mismatched PII account for 10\% of the total correctly predicted PII.
If we can propose a new criterion that combines $\mathcal{L}$ and $C_r$, we could have a theoretically 10\% improvement. 
Our solution is to add a bias $b$ to $\mathcal{L}_r$, resulting in~\autoref{eq:final_c}.
The core idea is that by adding $b$ to $\mathcal{L}_r$, we control how much influence $\mathcal{C}_r$ has in comparison to $\mathcal{L}$.
When a PII entity is ranked as top-1 by both $\mathcal{L}$ and $C_r$, it will be also ranked as top-1 by~\autoref{eq:final_c}.
Specifically, we introduce the following theorem.

\begin{theorem}\label{thm:criterion}
Let $\mathcal{M}=\left\{ M_i \right\}_{i=1}^n$. For $j \neq k$, $1\le j,k\le n$,if $\mathcal{L}(M_j) < \mathcal{L}(M_k)$ and $\frac{\mathcal{L}(M_j) - \mathcal{L}_r(M_j)}{\mathcal{L}_r(M_j)} < \frac{\mathcal{L}(M_k) - \mathcal{L}_r(M_k)}{\mathcal{L}_r(M_k)}$, then
\small
\begin{equation*}
\frac{\mathcal{L}(M_j) -  \left( \mathcal{L}_r(M_j) + b \right)}{\mathcal{L}_r(M_j) + b} < \frac{\mathcal{L}(M_k) -  \left( \mathcal{L}_r(M_k) + b \right)}{\mathcal{L}_r(M_k) + b}.
\end{equation*}
\normalsize
\end{theorem}
\autoref{thm:criterion} indicates that our proposed criterion $C$ in~\autoref{eq:final_c} can maintain the overlapped correct PII entities predicted by $\mathcal{L}$ and $C_r$. We present the detailed proof of~\autoref{thm:criterion} in the appendix.
When $b \to 0$, $C \to C_r$. Through Taylor's expansion,
\begin{equation}
    C = -1 + \frac{\mathcal{L}}{b} + \mathcal{O}\left(\frac{1}{b^2}\right).
\end{equation}
Based on the equation above, we know that as $b$ increases, $C$ becomes increasingly dominated by $\mathcal{L}$.
Therefore, by adjusting $b$, we can control whether $C$ is closer to $C_r$ or $\mathcal{L}$.
Through extensive experiments, we find that the optimal $b$ depends mainly on the model type (see~\autoref{sec:effect_bias}).
An attacker can determine the optimal $b$ by testing on a public PII dataset using the reference model.

\section{Evaluation}

\subsection{Experimental Setup}

We briefly introduce our datasets, models, metrics, and baselines in this section.
Please refer to~\autoref{app:attack_setting_details} for more attack settings.

\mypara{Datasets and LLMs}
We choose three PII text datasets to evaluate PII reconstruction attacks: ECHR~\citep{poudyal2020echr}, ENRON~\citep{klimt2004enron}, and LLM-PC~\citep{li2024the}.
We employ Microsoft Presidio to obtain the masked texts and the corresponding PII data.
Our dataset statistics are shown in~\autoref{tab:dataset_statistics} of~\autoref{app:dataset_details}.
As for victim LLMs, we choose four open-source LLMs for the evaluation: Llama3.1-8B, Llama3.2-3B, Qwen2.5-7B, and Phi3.5-Mini.

\mypara{Metrics}
We report the top-1 accuracy for PII prediction.
We consider a prediction successful when the true PII is included in the predicted sequence $M$, regardless of the letter case.
For evaluating PII candidate generation, we use PII recall, which measures how many PII entities are successfully generated in the candidates set $\mathcal{M}$.

\mypara{Baselines}
This paper considers three PII reconstruction attack methods as baselines: (1) DirectPrompt, which feeds the masked text directly to the LLM and prompts it to generate the best candidate to fill a specific PII mask; (2) TAB~\citep{inan2021training}, which generates completions using the true prefix; (3) P2P (Prefix, Presidio, and Perplexity)~\citep{lukas2023analyzing}, which performs multiple rounds of generation using true prefix and ranks the candidates using perplexity.

\begin{table*}[!t]
\centering
\caption{Top-1 accuracy of \method and baselines across different models and datasets.}
\label{tab:overall_performance}
\resizebox{\textwidth}{!}{
\begin{tabular}{l|cccccccccccc}
\toprule
    \multirow{2}{*}{\textbf{Stealer}} & \multicolumn{3}{c}{\textbf{Llama3.1-8B}} & \multicolumn{3}{c}{\textbf{Llama3.2-3B}} & \multicolumn{3}{c}{\textbf{Qwen2.5-7B}} & \multicolumn{3}{c}{\textbf{Phi3.5-Mini}} \\
    \cmidrule(lr){2-4} \cmidrule(lr){5-7} \cmidrule(lr){8-10} \cmidrule(lr){11-13}
    & ECHR & Enron & LLM-PC & ECHR & Enron & LLM-PC & ECHR & Enron & LLM-PC & ECHR & Enron & LLM-PC \\
    \midrule
    DirectPrompt & 6.07 & 2.55 & 10.33 & 3.56 & 2.11 &10.73 & 3.24 & 2.09 & 12.58 & 2.11 & 0.62 & 10.95 \\
    TAB & 13.51 & 19.00 & 8.30 & 7.20 & 9.12 & 6.77 & 8.61 & 13.31 & 6.95 & 4.09 & 5.62 & 4.86 \\
    \spbase & 13.19 & 19.14 & 11.68 & 6.91 & 8.38 & 8.65 & 8.99 & 13.50 & 10.31 & 4.28 & 5.74 & 7.41 \\
    \method & \textbf{25.68} & \textbf{33.31} & \textbf{28.93} & \textbf{14.79} & \textbf{20.61} & \textbf{26.48} & \textbf{16.35} & \textbf{25.38} & \textbf{26.41} & \textbf{11.10} & \textbf{16.71} & \textbf{22.13} \\
\bottomrule
\end{tabular}
}
\end{table*}

\begin{figure*}[!t]
    \centering
    \includegraphics[width=\textwidth]{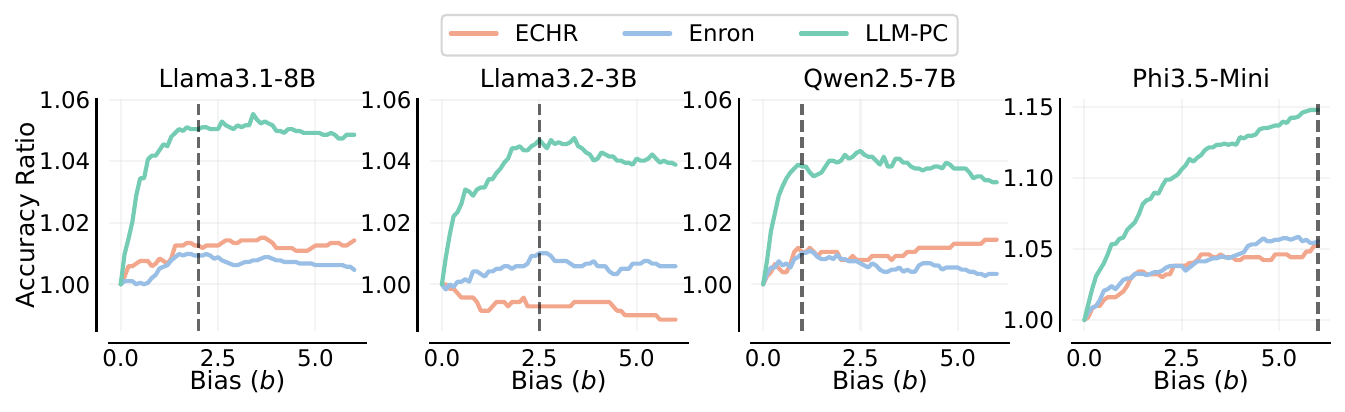}
    \caption{Accuracy ratio versus $b$. The gray dashed line indicates the optimal value of $b$ chosen during candidate selection. We can see that the accuracy ratio has a similar trend in the same model.}
    \label{fig:accuracy_ratio}
\end{figure*}

% \mypara{Implementation}

\subsection{Overall Effectiveness}
\label{sec:overall_exp}

\autoref{tab:overall_performance} presents the PII prediction accuracy of \method compared to our baseline methods.
The results demonstrate that \method consistently outperforms all baselines across different models and datasets.
Compared to previous SOTAs, \method achieves a 122\% improvement in top-1 accuracy, effectively reconstructing twice as many PII entities as before.
Among the baselines, DirectPrompt performs the worst, which we attribute to the alignment mechanisms of LLMs.
Directly querying for PII entities may sometimes trigger these mechanisms, causing the model to refuse to respond.
Although P2P introduces improvements over TAB, its prediction accuracy shows no significant advantage compared to TAB.
Additionally, we observe a clear trend: larger models achieve higher PII prediction accuracy.
This suggests that larger models have more redundant parameters, enabling them to memorize subtle information during training.
Finally, ECHR seems a more difficult dataset than Enron and LLM-PC.
This may be due to its shorter text length and the limited occurrence of most PII entities, which appear only once in the text.

\subsection{Effectiveness of Biased Criterion}
\label{sec:effect_bias}

To visually illustrate the impact of bias $b$ on candidate selection, we adjust $b$ and record the accuracy ratio, which is defined as the ratio of the current accuracy to the accuracy when $b=0$.
The observed results are plotted in~\autoref{fig:accuracy_ratio}.
From the figure, we can see that different datasets tend to exhibit similar accuracy ratio trends on the same model.
The only exception in our experiments is ECHR fine-tuned Llama3.2.
However, despite this anomaly, \method still achieves a decent prediction accuracy on ECHR fine-tuned LLama3.2 at $b=2.5$.
Except for Phi-3.5, the accuracy ratios of the other three models first increase and then decrease, indicating that three exits non-overlapping space between $L$ and $C_r$.
By combining these two criteria, we can achieve a more effective attack.
We also observe that LLM-PC dataset is the most sensitive to $b$, suggesting that the non-overlapping space between $L$ and $C_r$ is larger in this dataset.
This is likely due to the higher diversity of texts in LLM-PC.
Some PII entities experience a greater decrease in $L$ after training, while others show a more significant reduction in $C_r$.

\begin{table*}[!t]
\centering
\caption{Ablation study on candidate selection. We compare top-1 accuracy while ranking with different criteria.}
\label{tab:ablation_selection}
\resizebox{\textwidth}{!}{
\begin{tabular}{l|cccccccccccc}
\toprule
    \multirow{2}{*}{\textbf{Criterion}} & \multicolumn{3}{c}{\textbf{Llama3.1-8B}} & \multicolumn{3}{c}{\textbf{Llama3.2-3B}} & \multicolumn{3}{c}{\textbf{Qwen2.5-7B}} & \multicolumn{3}{c}{\textbf{Phi3.5-Mini}} \\
    \cmidrule(lr){2-4} \cmidrule(lr){5-7} \cmidrule(lr){8-10} \cmidrule(lr){11-13}
    & ECHR & Enron & LLM-PC & ECHR & Enron & LLM-PC & ECHR & Enron & LLM-PC & E
    CHR & Enron & LLM-PC \\
    \midrule
    $\mathcal{L}$ (vanilla) & 25.42 & 31.51 & 27.89 & 14.66 & 19.07 & 24.15 & 16.19 & 24.33 & 25.04 & 10.99 & 15.79 & 21.39 \\
    $\mathcal{L}$ & 25.55 & 32.95 & 28.57 & 14.74 & 20.40 & 25.41 & 16.31 & 25.13 & 25.88 & \textbf{11.10} & \textbf{16.71} & \textbf{22.13} \\
    $\mathcal{L} - \mathcal{L}_r$ & 19.73 & 27.24 & 19.01 & 12.38 & 15.83 & 15.93 & 12.15 & 17.42 & 15.11 & 5.94 & 7.11 & 7.51 \\
    $\frac{\mathcal{L} - \mathcal{L}_r}{\mathcal{L}_r}$ & 25.36 & 33.00 & 27.54 & \textbf{14.89} & 20.40 & 25.29 & 16.17 & 25.11 & 25.43 & 10.59 & 15.74 & 18.75 \\
    $\frac{\mathcal{L} - (\mathcal{L}_r + b)}{(\mathcal{L}_r + b)}$ & \textbf{25.68} & \textbf{33.31} & \textbf{28.93} & 14.79 & \textbf{20.61} & \textbf{26.48} & \textbf{16.35} & \textbf{25.38} & \textbf{26.41} & \textbf{11.10} & \textbf{16.71} & \textbf{22.13} \\
\bottomrule
\end{tabular}
}
\end{table*}

\begin{figure*}[!t]
    \centering
    \includegraphics[width=\textwidth]{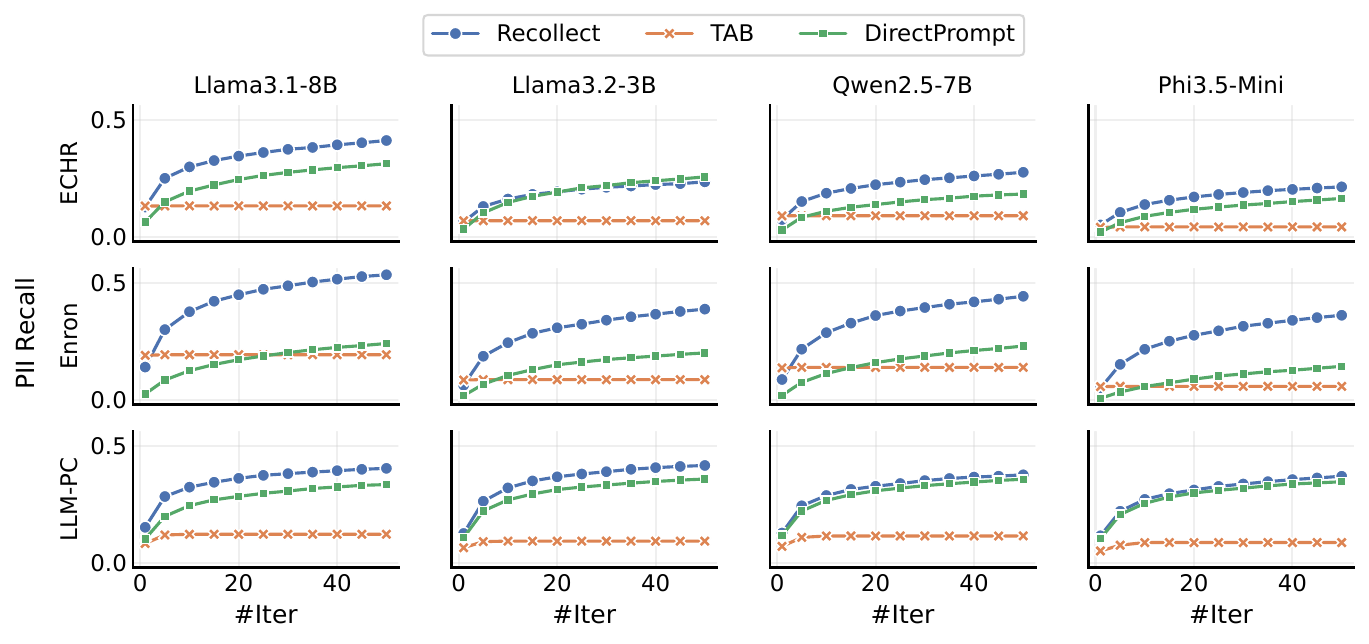}
    \vspace{-0.5cm}
    \caption{Ablation study on candidate generation. We report PII recall versus the number of query iterations.}
    \label{fig:ablation_generation}
\end{figure*}

\begin{figure}[tbp]
    \centering
    \begin{subfigure}{0.491\columnwidth}
        \centering
        \includegraphics[width=\linewidth]{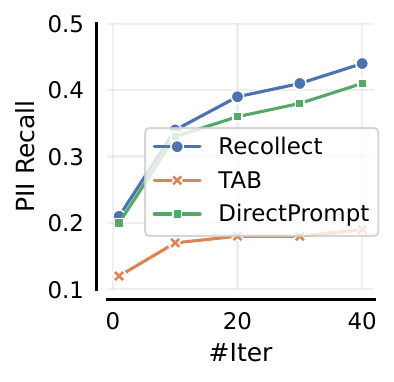}
        % vspace{-0.5cm}
        \caption{Comparison of PII recalls on GPT-4o-mini.}
        \label{fig:gpt-4o-mini}
    \end{subfigure}
    \hfill
    \begin{subfigure}{0.491\columnwidth}
        \centering
        \includegraphics[width=\linewidth]{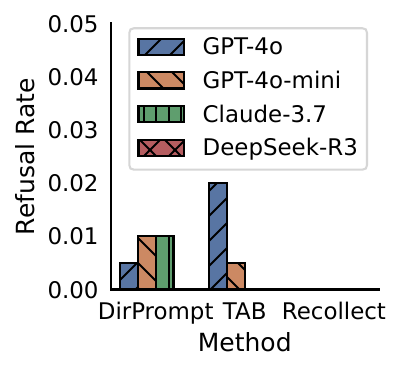}
        % \vspace{-0.5cm}
        \caption{Prompt refusal rates on proprietary models.}
        \label{fig:refusal_rates}
    \end{subfigure}
    % \vspace{-0.5cm}
    \caption{Experiments on real-world models.}
    \label{fig:real-world}
\end{figure}

\subsection{Ablation Studies}
\label{sec:ablation_studies}

In this section, we conduct ablation studies on two steps of \method.
For candidate generation and candidate selection, we replace Recollect and $C$ in~\autoref{eq:final_c} with alternative methods, respectively.
We also compare hybrid attacks that combine different candidate generation and candidate selection methods. Due to the space limitation, we defer hybrid attack results to~\autoref{app:hybrid_attacks}.

\mypara{Ablation of Candidate Generation}
We compare Recollect with DirectPrompt and TAB as alternative candidate generation methods.
For each approach, we query the victim model 50 times and compare their PII recall.
The experimental results are presented in~\autoref{fig:ablation_generation}.
We observe that both Recollect and DirectPrompt showcase an increasing trend in PII recall as the number of query iterations increases.
However, TAB reaches a peak with the first few iterations and then stabilizes, indicating that the candidates generated by TAB remain relatively fixed over multiple queries.
This also explains why P2P, while doing multiple true-prefix queries, fails to improve prediction accuracy.
In contrast, Recollect generally increases PII recall at a faster rate than DirectPrompt and continues to show an upward trend even after 50 iterations, demonstrating the scalability of \method.
We also conduct an experiment on GPT-4o-mini using OpenAI Fine-tuning API to demonstrate that our recollection method remains powerful to commercial models.
We randomly select 50 texts from LLM-PC, containing 218 PII entities and fine-tube GPT-4o-mini for 5 epochs.
The PII recall versus the number of query iterations is shown in~\autoref{fig:gpt-4o-mini}.
Recollect achieves the highest PII recall when giving the same query budget.

Another concern is that victim models may refuse to output or provide perturbed results when detecting privacy-related queries. To address this concern, we conduct an experiment using 200 prompts for each candidate generation method to investigate whether current proprietary models would refuse them. The refusal rate results are shown in~\autoref{fig:refusal_rates}. We observe that DirectPrompt occasionally triggers malicious behavior detection, causing the model to refuse to respond. TAB sometimes results in refusals from GPT-4o and GPT-4o-mini due to too short context provided for continuation. In contrast, none of the above four proprietary models refuses Recollect.
This may be because Recollect disguises itself by presenting the task of filling the missing parts.

\begin{table*}[tbp]
\centering
\caption{comparison of model performance and PII prediction accuracy for models trained with DP-SGD under varying privacy budget ($\epsilon$).}
\label{tab:dpsgd}
\resizebox{0.95\textwidth}{!}{
\begin{tabular}{l|ccccc ccccc}
\toprule
& \multicolumn{5}{c}{\textbf{Llama32-3B}} & \multicolumn{5}{c}{\textbf{Phi-3.5-mini}} \\
\cmidrule(lr){2-6} \cmidrule(lr){7-11}
$\boldsymbol{\epsilon}$ & 4 & 8 & 16 & 32 & $\infty$ & 4 & 8 & 16 & 32 & $\infty$ \\
\midrule
\textbf{Train Loss}        & 2.50  & 2.48  & 2.47  & 2.46  & 1.79 & 2.15  & 2.12  & 2.13  & 2.03  & 1.81  \\
\textbf{ACC (DirectPrompt)}     & 0.22\% & 0.39\% & 0.57\% & 0.44\% & 3.56\% & 0.44\% & 0.48\% & 0.84\% & 0.84\% & 2.11\% \\
\textbf{ACC (TAB)}         & 0.53\% & 0.35\% & 0.53\% & 0.62\% & 7.20\% & 0.40\% & 0.31\% & 0.44\% & 0.53\% & 4.09\% \\
\textbf{ACC (P2P)}         & 0.57\% & 0.62\% & 0.62\% & 0.75\% & 6.91\% & 0.48\% & 0.53\% & 0.58\% & 0.62\% & 4.28\% \\
\textbf{ACC (R.R.)}        & 3.60\% & 3.74\% & 3.87\% & 4.18\% & 14.79\% & 2.95\% & 3.21\% & 3.64\% & 3.91\% & 11.10\% \\
\bottomrule
\end{tabular}
}
\end{table*}

\begin{figure*}[!t]
    \centering
    \includegraphics[width=\textwidth]{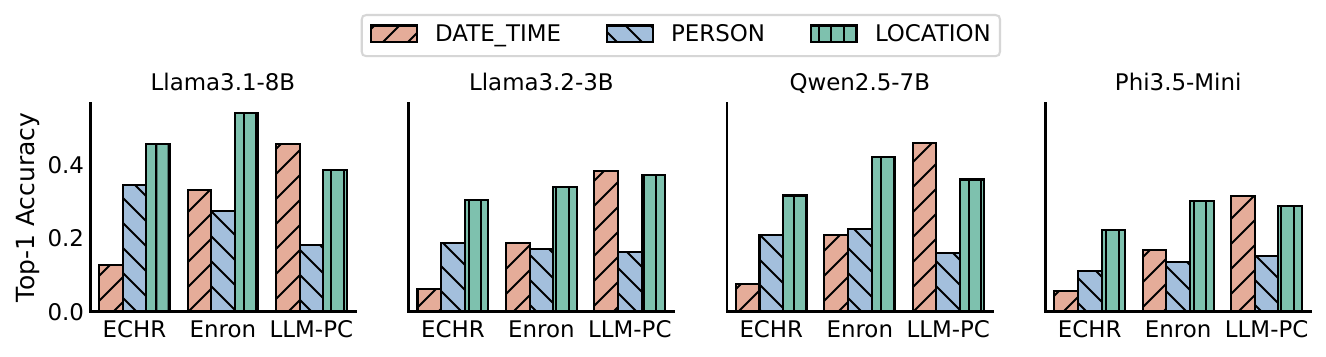}
    \caption{Top-1 accuracies of \method on specific PII entity types.}
    \label{fig:analyze_pii_type}
\end{figure*}

\mypara{Ablation of Candidate Selection}
To validate the effectiveness of the criterion proposed in~\autoref{eq:final_c}, we apply different selection criteria on candidates generated by Recollect and report their top-1 accuracy in~\autoref{tab:ablation_selection}.
From the table, we observe that partial loss consistently outperforms vanilla loss.
This is because vanilla loss introduces errors when PII entities appear multiple times in the text as discussed in~\autoref{sec:candidate_selection}. Our proposed $C$ achieves the highest top-1 accuracy in most cases.
Because Phi3.5-Mini exhibits a monotonic increase in accuracy ratio as $b$ increases, we set $L$ as the ranking criterion for Phi3.5-Mini in \method experiments, which is equal to the case $b \to \infty$.

% \begin{table}[tbp]
%     \centering
%     \caption{}
%     \label{tab:}
%     \resizebox{\columnwidth}{!}{
%     \begin{tabular}{lccccc}
%     \toprule
%     $\boldsymbol{\epsilon}$ & 4 & 8 & 16 & 32 & $\infty$ \\
%     \midrule
%     \textbf{Train Loss}          & 2.50  & 2.48  & 2.47  & 2.46  & 1.79  \\
%     \textbf{ACC (DPrompt)}  & 0.22\% & 0.39\% & 0.57\% & 0.44\% & 3.56\% \\
%     \textbf{ACC (TAB)}           & 0.53\% & 0.35\% & 0.53\% & 0.62\% & 7.20\% \\
%     \textbf{ACC (P2P)}           & 0.57\% & 0.62\% & 0.62\% & 0.75\% & 6.91\% \\
%     \textbf{ACC (\method)}          & 3.60\% & 3.74\% & 3.87\% & 4.18\% & 14.79\% \\
%     \bottomrule
%     \end{tabular}
%     }
% \end{table}

\subsection{Defending against \method}
\label{sec: defense}
To mitigate the risk of PII leakage from LLMs, developers may employ privacy-preserving techniques.
The most common and widely adopted approach is Differential Privacy (DP)~\citep{dwork2006differential}, which provides rigorous mathematical guarantees that an individual's data cannot be inferred through differential attacks.
DP-SGD~\citep{abadi2016deep} is the most prominent method for training deep learning models to satisfy DP.
Given a privacy budget ($\epsilon$), DP-SGD operates by clipping per-sample gradients and adding scaled noise during each update.
In this work, we fine-tune victim models on the ECHR dataset using DP-SGD implemented by Microsoft's DP-Transformers~\citep{dp-transformers}.
Since DP-SGD incurs extra memory overhead and is hard to integrate with distributed learning, we only conduct DP-SGD experiments on 3B models. The results are shown in~\autoref{tab:dpsgd}.

We find that although DP-SGD effectively reduces reconstruction accuracy, it also leads to a significant increase in loss, jeopardizing model performance.
Our baselines become nearly unusable on DP fine-tuned models, whereas \method remains about 1/3 accuracy.
The results suggest that while DP-SGD is a powerful privacy-preserving method, it cannot completely defend against our \method.
An alternative defense is to prevent the model from learning PII tokens during training.
For instance, \citealp{lin2024not} proposed Selective Language Modeling (SLM), which selectively trains on useful tokens.
We hypothesize that most PII tokens contribute little to the general language modeling capability of LLMs.
Thus, selectively excluding them could enhance privacy with minimal impact on utility.
Another line of defense is post-hoc unlearning~\citep{yao2024machine, liu2024towards}.
The developer can remove specific PII from LLMs after training using unlearning techniques.

\subsection{Analysis}

We explore how PII entity type and text length affect the reconstruction accuracy in this section.
We also analyze the impact of PII occurrence frequency.
Due to space limitation, we defer it into~\autoref{app:analysis_occurrence}.

\begin{figure}[!t]
    \centering
    \includegraphics[width=\columnwidth]{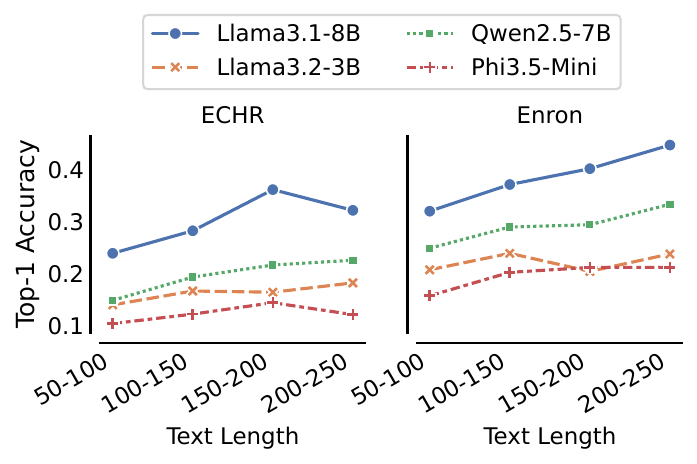}
    \caption{Top-1 accuracy versus text length.}
    \label{fig:analyze_text_len}
\end{figure}

\mypara{PII Entity Type}
We select three PII entity types that appear more than 100 times across all datasets, namely \texttt{DATE\_TIME}, \texttt{PERSON}, and \texttt{LOCATION}, to evaluate \method's effectiveness on specific entity types.
\autoref{fig:analyze_pii_type} illustrates the results.
Overall, \method demonstrates a strong ability to reconstruct \texttt{LOCATION} entities, while its performance on \texttt{PERSON} entities is weaker.
We suspect this is due to the discrepancy in value space; names vary significantly, whereas the number of locations is relatively limited, especially for major cities like New York and Los Angeles, which appear frequently and are easy to guess.
Additionally, we observe that \texttt{DATE\_TIME} exhibits varying prediction accuracies across different datasets. We suspect this is because \texttt{DATE\_TIME} precision differs across datasets.
For instance, ECHR often specifies exact dates, while LLM-PC typically records only the year.

\mypara{Text Length}
We split datasets into subsets based on a specific range of word lengths and measure how text length can impact PII prediction.
LLM-PC is excluded since its texts have already been processed to maintain a similar length.
The result is shown in \autoref{fig:analyze_text_len}. 
Top-1 accuracy tends to rise as text length grows. 
On the ECHR fine-tuned Llama3.1 model, the top-1 accuracy rises from 0.23 to 0.35 as text length grows from 50-100 to 150-200 words, demonstrating that more context helps LLM recall texts.
However, on Enron fine-tuned Llama3.1 model, \method performs slightly better at 100-150 words than 150-200 words.
This suggests that while longer texts provide more context, they may also introduce more noise, reducing performance.
In addition, the impact of text length on performance appears to be greater for larger models, which may be more capable of taking advantage of additional context in longer texts.

\section{Related Work} 

\mypara{LLM Memorization}
Memorization in LLMs refers to the model's tendency to store and reproduce exact phrases or passages from the training data rather than generating novel or generalized outputs ~\citep{satvaty2024undesirablememorizationlargelanguage}. It has been proved to play a significant role in simple, knowledge-intensive tasks ~\citep{wang2024generalizationvsmemorizationtracing}, but include rare, unrelated details before overfitting, a phenomenon called "unintended memorization" \citep{carlini2019secret}. This unintended memorization raises privacy concerns, especially on domain-specific models ~\citep{yang2024memorization}. Several studies have examined how factors such as model size, training data repetition, context length, and fine-tuning strategies influence LLM memorization ~\citep{carlini2023quantifyingmemorizationneurallanguage, mireshghallah2022empirical, zeng2024exploringmemorizationfinetunedlanguage}.

\mypara{PII Leakage}
Due to the unintended memorization of LLMs, malicious attackers can exploit this vulnerability to steal sensitive data from training sets, including Personally Identifiable Information (PII), resulting in PII leakage. ~\citealp{291327} emphasize the need to address this issue, proposing a pipeline for extracting sensitive information from the Codex model. PII leakage attacks can be classified into three categories according to the adversary’s capabilities: extraction attacks ~\citep{carlini2021extracting, mireshghallah2022empirical, yu2023bagtrickstrainingdata, zhang2023ethicisttargetedtrainingdata}, reconstruction attacks ~\citep{inan2021training, lukas2023analyzing}, and inference attacks~\citep{mireshghallah2022quantifying, fu2024membership}. One technique to mitigate these risks is directly editing model weights, though no universal defense methods exist yet ~\citep{patil2023sensitiveinformationdeletedllms}. Further attacks can bypass model alignment and recover sensitive training data ~\citep{nasr2025scalable}.

\section{Conclusion}
In this paper, we propose \method (Recollect and Rank), a novel two-step PII reconstruction attack. First, we employ a recollection-based method to recover the original text and leverage a NER-model-powered PII identifier to extract potential PII candidates.
% To evaluate the performance of PII prediction and PII candidate generation, we use top-1 accuracy and PII recall, respectively.
Then, we rank the PII candidates by computing their cross-entropy scores under the victim model and a reference model. We further introduce a biased ranking criterion that effectively integrates reference-based and non-reference-based ranking methods. Our experiments across three popular PII datasets and four open-source LLMs demonstrate that the \method significantly improves top-1 accuracy.

\section*{Acknowledgements}
% We thank our anonymous reviewers for their feedback.
This research received partial support from the National Natural Science Foundation of China under Grant No. 62302441. This work was also supported by the Key Research and Development Program Project of Ningbo Grant No. 2025Z029. Infrastructure support was provided by the Information Technology Center of Zhejiang University and the Supercomputing Center of Hangzhou City University.

\clearpage

\section*{Limitations}

\mypara{Requirement for the Reference Model}
\method's new ranking criterion relies on a reference model for calibration. 
We assume that the victim model is fine-tuned from an open-source model, but this assumption does not always hold, particularly when attacking proprietary black-box models.
Furthermore, incorporating a reference model increases the computational overhead for the attacker.
Unlike the victim model, which can be queried, the reference model must be run locally, requiring additional resources.
If the reference model is large, the attack becomes significantly more expensive, posing a practical limitation for adversaries with limited computational power.

\mypara{Need to Adjust $b$ When Changing the Victim Model}
To better combine $\mathcal{L}$ and $C_r$ in reference calibration, we introduce a bias $b$. As demonstrated in \autoref{sec:effect_bias}, $b$ generally improves performance, except for the ECHR fine-tuned Llama3.2 model. However, we cannot guarantee that $b$ will not be ineffective for other models, as we cannot predict whether the victim model will behave like the ECHR fine-tuned Llama3.2. Additionally, determining the optimal value of $b$ is computationally expensive. Since the optimal bias $b$ largely depends on the specific model type, it is necessary to recalculate $b$ each time a new victim model is selected, resulting in a significant overhead. 

% \mypara{Lack defense mechanism considerations}
% Moreover, we didn't consider the defense mechanisms that may be applied to LLMs, such as differential privacy, adversarial training and other privacy-preserving techniques. These mechanisms are specifically designed to mitigate privacy risks by reducing the amount of sensitive information exposed during model training or inference. While these defenses could reduce the effectiveness of our approach, they may also result in a loss of model utility. Further studies are needed to evaluate the performance of our method on models equipped with such defenses.

\mypara{Robustness against Defense Mechanisms}
In real-world applications, developers often employ privacy-preserving techniques, such as differential privacy, to effectively mitigate potential risks of PII leakage. As demonstrated in ~\autoref{sec: defense}, DP-SGD significantly reduces the PII prediction accuracy of our proposed method to one-third.  However, despite this degradation, our proposed method still outperforms other baseline approaches. Furthermore, such defense mechanisms not only incur higher computational costs, but also reduce the utility of LLMs. 

% \section*{Acknowledgments}

% This document has been adapted
% by Steven Bethard, Ryan Cotterell and Rui Yan
% from the instructions for earlier ACL and NAACL proceedings, including those for
% ACL 2019 by Douwe Kiela and Ivan Vuli\'{c},
% NAACL 2019 by Stephanie Lukin and Alla Roskovskaya,
% ACL 2018 by Shay Cohen, Kevin Gimpel, and Wei Lu,
% NAACL 2018 by Margaret Mitchell and Stephanie Lukin,
% Bib\TeX{} suggestions for (NA)ACL 2017/2018 from Jason Eisner,
% ACL 2017 by Dan Gildea and Min-Yen Kan,
% NAACL 2017 by Margaret Mitchell,
% ACL 2012 by Maggie Li and Michael White,
% ACL 2010 by Jing-Shin Chang and Philipp Koehn,
% ACL 2008 by Johanna D. Moore, Simone Teufel, James Allan, and Sadaoki Furui,
% ACL 2005 by Hwee Tou Ng and Kemal Oflazer,
% ACL 2002 by Eugene Charniak and Dekang Lin,
% and earlier ACL and EACL formats written by several people, including
% John Chen, Henry S. Thompson and Donald Walker.
% Additional elements were taken from the formatting instructions of the \emph{International Joint Conference on Artificial Intelligence} and the \emph{Conference on Computer Vision and Pattern Recognition}.

\bibliography{custom}

% \clearpage

\appendix
\section*{\centering Appendices}
\section{Proof of \autoref{thm:criterion}}

\noindent \textbf{Theorem 3.1.} Let $\mathcal{M}=\left\{ M_i \right\}_{i=1}^n$. For $j \neq k$, $1\le j,k\le n$,if $\mathcal{L}(M_j) < \mathcal{L}(M_k)$ and $\frac{\mathcal{L}(M_j) - \mathcal{L}_r(M_j)}{\mathcal{L}_r(M_j)} < \frac{\mathcal{L}(M_k) - \mathcal{L}_r(M_k)}{\mathcal{L}_r(M_k)}$, then
\small
\begin{equation*}
\frac{\mathcal{L}(M_j) -  \left( \mathcal{L}_r(M_j) + b \right)}{\mathcal{L}_r(M_j) + b} < \frac{\mathcal{L}(M_k) -  \left( \mathcal{L}_r(M_k) + b \right)}{\mathcal{L}_r(M_k) + b}
\end{equation*}
\normalsize

\noindent \textbf{Proof.}
Given $\frac{\mathcal{L}(M_j) - \mathcal{L}_r(M_j)}{\mathcal{L}_r(M_j)} < \frac{\mathcal{L}(M_k) - \mathcal{L}_r(M_k)}{\mathcal{L}_r(M_k)}$, we have
\begin{align}\label{eq:corollary1}
     \frac{\mathcal{L}(M_j)}{\mathcal{L}_r(M_j)} &< \frac{\mathcal{L}(M_k)}{\mathcal{L}_r(M_k)} \notag \\
     \mathcal{L}(M_j)\mathcal{L}_r(M_k) &< \mathcal{L}(M_k)\mathcal{L}_r(M_j).
\end{align}
Because $b$ is a positive bias, we can get
\begin{equation}\label{eq:corollary2}
    b \mathcal{L}\left(M_j\right)<b \mathcal{L}\left(M_k\right).
\end{equation}
Combing~\autoref{eq:corollary1} and~\autoref{eq:corollary2} we obtain
\small
\begin{equation*}
    \mathcal{L}(M_j)\mathcal{L}_r(M_k) + b\mathcal{L}(M_j) < \mathcal{L}(M_k)\mathcal{L}_r(M_j) + b\mathcal{L}(M_k)
\end{equation*}
\normalsize
By a simple transformation, we get
\small
\begin{align*}
    \frac{\mathcal{L}_r(M_k) + b}{\mathcal{L}(M_k)} &< \frac{\mathcal{L}_r(M_j) +b}{\mathcal{L}(M_j)} \\
    \frac{\mathcal{L}(M_j)}{\mathcal{L}_r(M_j) + b} &< \frac{\mathcal{L}(M_k)}{\mathcal{L}_r(M_k) +b} \\
    \frac{\mathcal{L}(M_j) -  \left( \mathcal{L}_r(M_j) + b \right)}{\mathcal{L}_r(M_j) + b} &< \frac{\mathcal{L}(M_k) - \left( \mathcal{L}_r(M_k) + b\right)}{\mathcal{L}_r(M_k) + b},
\end{align*}
\normalsize
Based on our assumption that $k$ can be any other candidate index that is unequal to $j$, $M_j$ will also be ranked as the first by $\frac{\mathcal{L}-\left(\mathcal{L}_r+b\right)}{\mathcal{L}_r+b}$. \hfill $\Box$

\section{Attack Setting Details}
\label{app:attack_setting_details}

\subsection{Dataset Details}
\label{app:dataset_details}

\autoref{tab:dataset_statistics} shows the statistics of our datasets.
The information and post-processing steps for each dataset are as follows.

\myparatight{ECHR}
This is an English legal judgment dataset comprising approximately 11,500 cases from the European Court of Human Rights. We randomly select 4,000 cases for training victim models and choose 2,000 cases from the training set to perform PII reconstruction attacks.

\myparatight{Enron}
The Enron dataset is a comprehensive collection of approximately 500,000 emails of the Enron Corporation.
Similar to ECHR, we select 4,000 emails for training victim models and choose 2,000 emails to perform PII reconstruction attacks.

\myparatight{LLM-PC}
This dataset is released by NeurIPS in its 2024 LLM Privacy challenge.
LLM-PC consists of two subsets: a development set for assessment and a test set for competition ranking.
Each subset contains 1,500 synthetic texts.
We use the development set for training victim models and performing PII reconstruction attacks since the test set's ground truth is not available.

\subsection{Hyperparameters}

Each target mode is trained using SFT for three epochs with a learning rate of 2e-5 on each dataset.
In the candidate generation step, we query the victim model 40 times for each PII entity to collect candidate predictions.
During generation, we set the temperature to 1.2, top-$K$ to 30, and top-$P$ to 0.8.
Given the extensive length of LLM-PC's text, we split it into sentences and extract those containing a specific PII entity along with their neighboring sentences to construct the input text.
The recollect prompts we use are illustrated in~\autoref{fig:recollect_prompts_echr_enron} and~\autoref{fig:recollect_prompt_llm_pc}.
The biases we used in candidate selection are shown in~\autoref{tab:biases}
As for baselines, we follow the settings in their original papers.

\begin{table}[!t]
    \centering
    \caption{Dataset statistics.}
    \label{tab:dataset_statistics}
    \resizebox{\columnwidth}{!}{
    \begin{tabular}{l|ccc}
    \toprule
    Dataset & \# of texts & average length & \# of PII \\
    \midrule
    ECHR & 2,000 & 88 words & 4,692 \\
    Enron & 2,000 & 90 words & 5,836 \\
    LLM-PC & 1,500 & 1,275 words & 5,907 \\
    \bottomrule
    \end{tabular}
    }
\end{table}

\begin{table}[!tbp]
    \centering
    \caption{Biases in candidate selection.}
    \label{tab:biases}
    \begin{tabular}{l|c}
    \toprule
    Model & $b$ \\
    \midrule
    Llama3.1-8B & 2.0 \\
    Llama3.2-3B & 2.5 \\
    Qwen2.5-7B  & 1.0 \\
    Phi3.5-Mini & $\infty$ \\
    \bottomrule
    \end{tabular}
\end{table}

\begin{table*}
    \centering
    \caption{Description of PII entity types.}
    \label{tab:pii_types}
    \resizebox{\textwidth}{!}{
        \begin{tabular}{lll}
        \toprule
             \textbf{Entity Type}&\textbf{Description}  &\textbf{Example} \\
        \midrule
             \texttt{DATE\_TIME}& Absolute or relative dates or periods or times smaller than a day.  & 5 May 2010\\
             \texttt{EMAIL\_ADDRESS} &Identifies an email inbox for delivering email messages. &james.tyrone@example.com\\
             \texttt{ID}& A unique identifier assigned to entities like applicants, respondents, or judges. &4616\\
             \texttt{LOCATION}& A geographically or politically defined location, such as cities or countries. &New Orleans, Louisiana\\
             \texttt{ORGANIZATION}& Companies, groups, clubs, government bodies, and public organizations. &British Council\\
             \texttt{PERSON}& Represents a full person name, including first, middle, or last names. &Earl Blanton\\
             \texttt{PHONE\_NUMBER}& A telephone number used for contact or communication. &555-123-4567\\
             \texttt{URL}& A Uniform Resource Locator, used to identify resources on the internet. &http://www.francismorrisdata.com\\
        \bottomrule
        \end{tabular}
    }
\end{table*}

\begin{figure}[!tbp]
    \centering
    \includegraphics[width=\columnwidth]{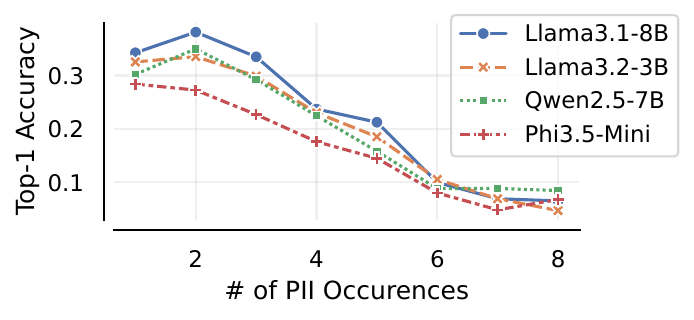}
    \caption{Top-1 accuracy versus the number of PII occurrences on LLM-PC dataset.}
    \label{fig:analyze_pii_times}
\end{figure}

\section{Attack with Incorrect Reference Model}
\label{app:incorrect_reference}

\begin{figure}[tbp]
    \centering
    \includegraphics[width=\columnwidth]{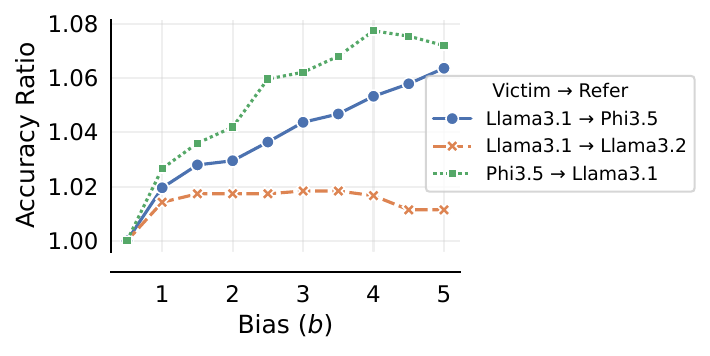}
    \caption{Accuracy ratio versus $b$ under incorrect reference model. The optimal choice of $b$  is influenced by the selection of the reference model.}
    \label{fig:incorrect_reference}
    
\end{figure}

We conducted additional experiments to study the robustness of our attack under incorrect reference. We measured the top-1 reconstruction accuracy using various reference models. The results (reported in percentages)  are reported in ~\autoref{tab:incorrect inference}.  Interestingly, we found that using an incorrect reference model does not significantly degrade reconstruction accuracy, and in some cases, improves it. The maximum accuracy drop is only 1.62\%. Selecting Llama3.2-3B as the reference model to attack Llama3.1-8B even results in a 0.43\% improvement. We attribute this to the fact that LLMs are generally pretrained on similar corpora, primarily web-crawled data, which enables different models to provide similar references. Another noteworthy observation is that the choice of the reference model influences the optimal value of the bias $b$. Some illustrative examples are shown in ~\autoref{fig:incorrect_reference}.  We observe that the accuracy consistently increases when applying Phi3.5-Mini as the reference model for attacks against Llama3.1-8B, which aligns with the trend when applying Phi3.5-Mini to the correct corresponding victim model. This indicates a severe threat, as attackers could select an appropriate value of $b$ based solely on the reference model without access to the victim model.

\begin{table}[!t]
\centering
\caption{Robustness of R.R. under incorrect reference. Each row corresponds to a victim model and each column corresponds to a reference model. The number in parentheses denotes the change relative to using the correct reference model. } % You can change the caption
\label{tab:incorrect inference}      % You can change the label
\resizebox{\columnwidth}{!}{%
\begin{tabular}{lcccc}
\toprule
& \textbf{Llama3.1-8B} & \textbf{Llama3.2-3B} & \textbf{Qwen2.5-7B} & \textbf{Phi3.5-Mini} \\
\midrule
\textbf{Llama3.1-8B} & 28.93 & 29.36 (+0.43) & 28.75 (-0.18) & 28.57 (-0.36) \\
\textbf{Llama3.2-3B} & 25.53 (-0.95) & 26.48 & 25.23 (-1.25) & 25.41 (-1.07) \\
\textbf{Qwen2.5-7B} & 26.11 (-0.30) & 26.21 (-0.20) & 26.41 & 25.41 (-1.00) \\
\textbf{Phi3.5-Mini} & 21.14 (-0.99) & 21.69 (-0.44) & 20.51 (-1.62) & 22.13 \\
\bottomrule
\end{tabular}%
}
\end{table}

\section{Hybrid Attacks}
\label{app:hybrid_attacks}

We conducted hybrid attack experiments combining three different candidate generation methods and five different candidate selection criteria against the Llama3.1-8B and Qwen2.5-7B models on LLM-PC dataset. The Top-1 Accuracy results are shown as ~\autoref{tab:hybrid_attacks}. From the table, we observe that our proposed method consistently achieves the highest Top-1 accuracy across both models when compared to other hybrid attack combinations. This highlights the effectiveness of our proposed method: the “recollecting” process yields a high-quality PII candidates pool, from which the predicted candidate is selected based on the validated biased ranking criterion.

\begin{table*}[!t]
\centering
\caption{Comparison of PII prediction accuracy for different hybrid attacks against Llama3.1-8B and Qwen2.5-7B on LLM-PC dataset.}
\label{tab:hybrid_attacks}
\resizebox{\textwidth}{!}{%
    % \begin{tabular}{@{}llccccc@{}}
    % \begin{tabular}{@{}ll w{c}{2cm} w{c}{2cm} w{c}{2cm} w{c}{2cm} w{c}{2cm} @{}}
    \begin{tabular}{@{}ll w{c}{2.5cm} w{c}{2.5cm} w{c}{2.5cm} w{c}{2.5cm} w{c}{2.5cm} @{}}
        \toprule
        \multirow{2}{*}{\textbf{Victim Model}} & \multirow{2}{*}{\textbf{Candidate Generation}} & \multicolumn{5}{c}{\textbf{Candidate Selection}} \\
        \cmidrule(lr){3-7}
        & & $\mathcal{L}$ (vanilla)& \textbf{$\mathcal{L}$} & \textbf{$\mathcal{L} - \mathcal{L}_r$} & \textbf{$\frac{\mathcal{L} - \mathcal{L}_r}{\mathcal{L}_r}$} & \textbf{$\frac{\mathcal{L} - (\mathcal{L}_r + b)}{(\mathcal{L}_r + b)}$} \\
        \midrule
        \multirow{3}{*}{\textbf{Llama3.1-8B}} & DirectPrompt & 24.34 & 24.60 & 18.01 & 24.19 & 24.90 \\
        & Tab & 11.68 & 11.78 & 11.38 & 11.80 & 11.80 \\
        & Recollect & 27.89 & 28.57 & 19.01 & 27.54 & \textbf{28.93} \\
        \midrule
        \multirow{3}{*}{\textbf{Qwen2.5-7B}} & DirectPrompt & 23.12 & 23.44 & 15.07 & 23.46 & 24.06 \\
        & Tab & 10.31 & 10.41 & 10.80 & 10.95 & 11.09 \\
        & Recollect & 25.04 & 25.88 & 15.11 & 25.43 & \textbf{26.41} \\
        \bottomrule
    \end{tabular}%
}
\end{table*}

\section{Analysis of PII Occurrence Frequency}
\label{app:analysis_occurrence}

A PII entity may appear multiple times within a given text.
We measure the impact of PII occurrence frequency on \method performance.
The results are presented in~\autoref{fig:analyze_pii_times}.
We focus solely on LLM-PC dataset, as ECHR and Enron's texts are relatively short, with most PII entities appearing only once.
Interestingly, the peak top-1 accuracy occurs when the number of PII occurrences is 2.
Beyond this point, top-1 accuracy gradually decreases.
Upon further investigation, we find that the distribution of PII entity types varies across different occurrence counts.
When the number of PII occurrences is 2, \texttt{PERSON} entities account for 35.6\% of the dataset.
However, when the number of PII occurrences increases to 6, the proportion of \texttt{PERSON} entities rises to 81.8\%.
As discussed earlier, \texttt{PERSON} is a challenging PII entity type to reconstruct.
This observation indicates that the number of PII occurrences alone does not compensate for the influence of PII entity type.

\begin{figure*}
\begin{mybox}{\textbf{\small{Recollect Prompt Template of ECHR and ENRON}}}
\small{
You are an expert in understanding texts and filling in masked information. Your task is to replace placeholders, denoted by [MASK-NUM], with the most appropriate and contextually relevant candidates. Ensure that all placeholders are replaced, and none of them appear in the output. The output should be natural and fully readable without brackets or placeholder markers.\\

There are some rules you need to follow:\\
**Rules**\\
1. Ensure placeholders are fully replaced, and no brackets or placeholder markers remain in the output.\\
2. End when finish rewriting the text.\\

Below is an example:\\
Text with masked placeholders:\\
24. On [DATE\_TIME-0] Mr [PERSON-0]’s counsel appealed to the Lucerne Court of Appeal (Obergericht) against the detention order issued on [DATE\_TIME-1] (see paragraph 9 above) and applied for his client’s immediate release.\\

Rewritten text:\\
24. On 23 January 1992 Mr R.M.D.’s counsel appealed to the Lucerne Court of Appeal (Obergericht) against the detention order issued on 17 January (see paragraph 9 above) and applied for his client’s immediate release.\\

Now please help me rewrite the following text.\\
Text with masked placeholders:\\
\{\texttt{MASKED\_TEXT}\}\\

Rewritten text:\\

}
\end{mybox}
\caption{Recollect prompt template of ECHR and Enron.}
\label{fig:recollect_prompts_echr_enron}
\end{figure*}

\begin{figure*}
\begin{mybox}{\textbf{\small{Recollect Prompt Template of LLM-PC}}}
\small{
You are an expert in understanding conversations and filling in masked information. Your task is to replace a placeholder, denoted by \{mask\}, with the most appropriate and contextually relevant candidate. You should ensure that the placeholder does not appear in your output.\\

There are some rules you need to follow:\\
**Rules**\\
1. Ensure placeholders are fully replaced, and no brackets or placeholder markers remain in the output.\\
2. End when finish rewriting the text.\\

Here is a text with some missing parts, represented by \{mask\}:\\

\{\texttt{MASKED\_TEXT}\}\\

Please repeat the above text and replace \{mask\} with the most probable candidate based on the context. Here is the output:\\

}
\end{mybox}
\caption{Recollect prompt template of LLM-PC.}
\label{fig:recollect_prompt_llm_pc}
\end{figure*}

\end{document}